\documentclass{article}

\usepackage{microtype}
\usepackage{graphicx}
\usepackage{subcaption}
\usepackage{booktabs} 
\usepackage{bm}

\usepackage{hyperref}

\usepackage[preprint]{icml2026}

\usepackage{algorithm}
\usepackage{algorithmic}
\usepackage{amsmath}
\usepackage{amssymb}
\usepackage{mathtools}
\usepackage{amsthm}
\usepackage{amsfonts}
\usepackage{float}
\usepackage[capitalize,noabbrev]{cleveref}

\theoremstyle{plain}

\theoremstyle{definition}

\theoremstyle{remark}

\usepackage[textsize=tiny]{todonotes}

\icmltitlerunning{Free-Form Density Estimation}

\begin{document}

\twocolumn[
\icmltitle{A CDF-First Framework for Free-Form Density Estimation}
  


  \icmlsetsymbol{equal}{*}

  \begin{icmlauthorlist}
    \icmlauthor{Chenglong Song}{equal,1}
    \icmlauthor{Mazharul Islam}{equal,2}
    \icmlauthor{Lin Wang}{1,3}
    \icmlauthor{Bing Chen}{4}
    \icmlauthor{Bo Yang}{3,1}
  \end{icmlauthorlist}
  \icmlaffiliation{1}{Shandong Key Laboratory of Ubiquitous Intelligent Computing, University of Jinan}
  \icmlaffiliation{2}{Shandong Provincial Key Laboratory of Green and Intelligent Building Materials, University of Jinan, Jinan, Shandong 250022, China}
  \icmlaffiliation{3}{Quancheng Shandong Laboratory, Jinan 250100, China}
  \icmlaffiliation{4}{David R. Cheriton School of Computer Science, University of Waterloo, Waterloo, Canada}

  \icmlcorrespondingauthor{Lin Wang}{wangplanet@gmail.com}

  \icmlkeywords{Conditional Density Estimation, Free-Form Densities, Cumulative Distribution Functions, Multivariate Density Estimation, Monotonic Networks}

   \vskip 0.3in
]





\printAffiliationsAndNotice{\icmlEqualContribution}

\begin{abstract}
Conditional density estimation (CDE) is a fundamental task in machine learning that aims to model the full conditional law \(\mathbb{P}(\mathbf{y} \mid \mathbf{x})\), beyond mere point prediction (e.g., mean, mode). A core challenge is \textit{free-form density estimation}—capturing distributions that exhibit multimodality, asymmetry, or topological complexity without restrictive assumptions. However, prevailing methods typically estimate the probability density function (PDF) directly, which is mathematically ill-posed: differentiating the empirical distribution amplifies random fluctuations inherent in finite datasets, necessitating strong \textit{inductive biases} that limit expressivity and fail when violated. We propose a CDF-first framework that circumvents this issue by estimating the cumulative distribution function (CDF), a stable and well-posed target, and then recovering the PDF via differentiation of the learned smooth CDF. Parameterizing the CDF with a Smooth Min-Max (SMM) network, our framework guarantees valid PDFs by construction, enables tractable approximate likelihood training, and preserves complex distributional shapes. For multivariate outputs, we use an autoregressive decomposition with SMM factors. Experiments demonstrate our approach outperforms state-of-the-art density estimators on a range of univariate and multivariate tasks.
\end{abstract}

\section{Introduction}\label{section1}
In supervised learning, point prediction—estimating the mean or mode of \(\mathbf{y}\) given \(\mathbf{x}\)—is often insufficient for high-stakes decision-making~\cite{NIPS2017_9ef2ed4b}. What is required instead is the full conditional law \(\mathbb{P}(\mathbf{y} \mid \mathbf{x})\), which enables calibrated uncertainty quantification, risk assessment, and probabilistic reasoning. This task is formalized as conditional density estimation: learning a measurable function \( p(\cdot \mid \mathbf{x}) : \mathbb{R}^{d_y} \to \mathbb{R}_{\geq 0} \) such that for each fixed \(\mathbf{x} \in \mathbb{R}^{d_x}\),
\begin{equation}
p(\mathbf{y} \mid \mathbf{x}) \geq 0, \quad \int_{\mathbb{R}^{d_y}} p(\mathbf{y} \mid \mathbf{x}) \, d\mathbf{y} = 1.   
\end{equation}

\begin{figure}
    \centering
    \includegraphics[width=\linewidth]{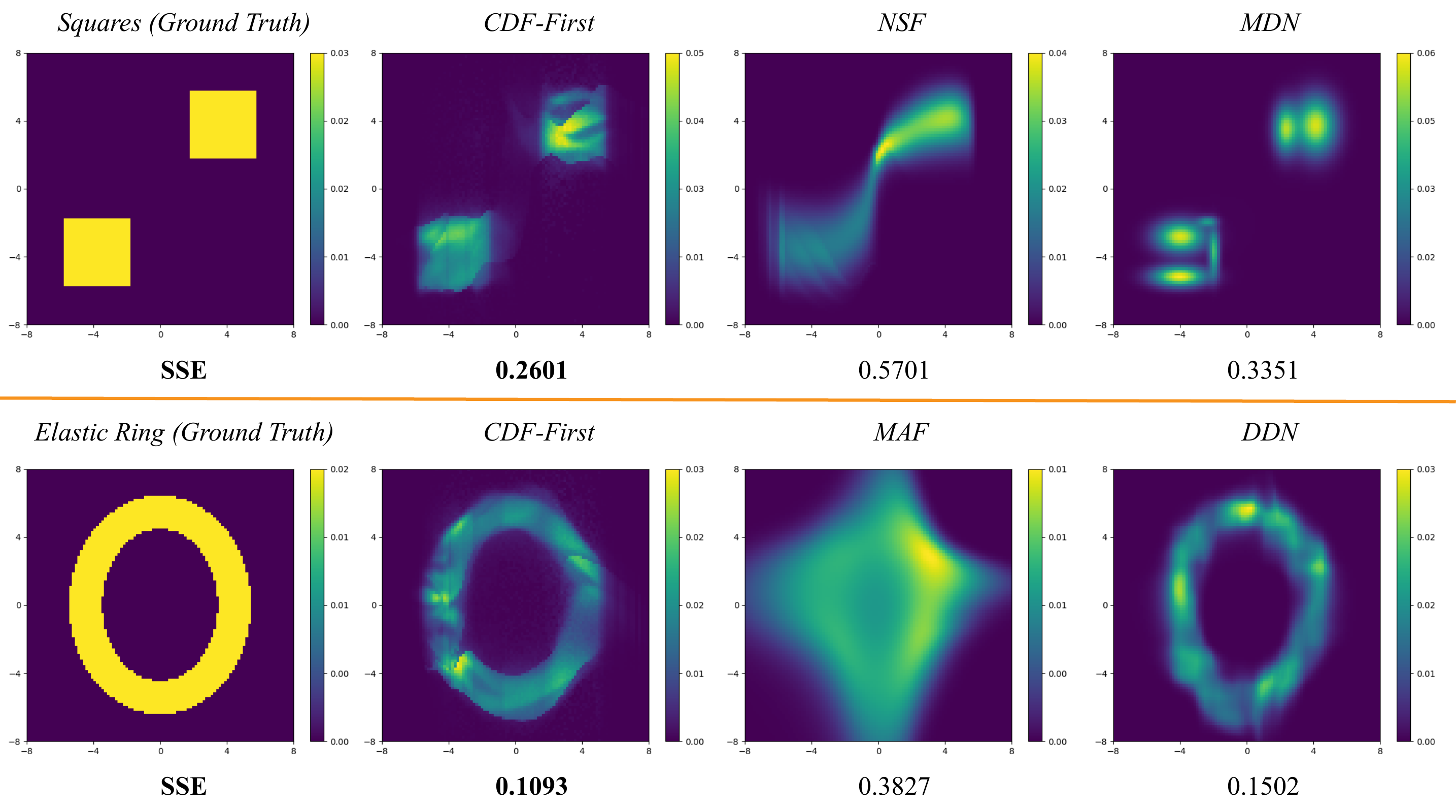}
    \caption{Comparison on toy tasks with complex support geometry: (\textit{top}) disconnected squares and (\textit{bottom}) ring-shaped densities. Our CDF-first model (\textit{second column}) accurately recovers the ground-truth distribution (\textit{leftmost column}), preserving sharp boundaries and topological holes. In contrast, MDN oversmooths multimodality, normalizing flows (NSF/MAF) fill topological voids, and DDN reduces spikiness but still blurs structural details. This demonstrates that modeling the CDF directly enables superior representation of free-form conditional densities.}
    \label{fig:structural_fidelity}
\end{figure}

However, real-world conditional distributions often exhibit complex characteristics that defy simple parametric forms. \textit{Free-form density estimation}: representing conditional distributions that may be multi-modal, skewed, heavy-tailed, or topologically complex (e.g., rings, disconnected components) without imposing restrictive parametric assumptions remains a fundamental challenge. This challenge is illustrated in Figure~\ref{fig:structural_fidelity}, where distributions with disconnected support or topological holes cannot be faithfully captured by standard methods. This leads to a core question: \textit{how can we model arbitrary distributional shapes without imposing restrictive assumptions that limit expressivity?}

Existing approaches to CDE typically estimate the probability density function directly through predefined generative mechanisms. Mixture density networks~\cite{citation:MDN} combine neural networks with parametric mixture models, but their fixed kernel families cannot adapt to arbitrary distributional shapes. Normalizing flows~\cite{citation:NF} transform simple base distributions through invertible mappings, but their invertibility requirement constrains the topological expressivity of the learned density. Discretization methods~\cite{citation:pixelcnn, Chen2021DeconvolutionalDN} partition the output space into bins, but they suffer from quantization artifacts and the curse of dimensionality. These assumptions limit expressivity and fail catastrophically when the true distribution violates them, as shown in Figure~\ref{fig:structural_fidelity} where MDN oversmooths multimodal structures, NFs fill topological holes, and discretization methods blur structural details.

The core difficulty stems from a mathematical reality: direct PDF estimation is ill-posed. The PDF is defined as the Radon-Nikodym derivative \(p = d\mu/d\mathbf{y}\) of the probability measure \( \mu \). Differentiation is an unbounded operator that amplifies high-frequency noise inherent in finite samples, making PDF estimation inherently unstable. This instability forces existing methods to incorporate strong regularization through their architectural assumptions, which inevitably restricts their ability to model free-form densities. In contrast, estimating the cumulative distribution function \(F(\mathbf{y} \mid \mathbf{x}) = \mu((-\infty, \mathbf{y}] \mid \mathbf{x})\) is well-posed: the mapping \( \mu \mapsto F \) is continuous under weak convergence, making CDF estimation stable with finite data.

We propose a paradigm shift: model the CDF, not its derivative. For a continuous distribution, the CDF \(F(\mathbf{y} \mid \mathbf{x}) = \int_{-\infty}^{\mathbf{y}} p(\mathbf{t} \mid \mathbf{x}) \, d\mathbf{t}\) is a stable, well-posed target: it is bounded in \([0,1]\), non-decreasing in each component of \(\mathbf{y}\), and robust to finite-sample perturbations because integration acts as a low-pass filter. If \(F(\cdot \mid \mathbf{x})\) is absolutely continuous, then by the Lebesgue differentiation theorem the PDF can be recovered via differentiation:
\begin{equation}
    p(\mathbf{y} \mid \mathbf{x}) = \frac{\partial^{d_y}}{\partial y_1 \cdots \partial y_{d_y}} F(\mathbf{y} \mid \mathbf{x}).
\end{equation}
This suggests reducing CDE to learning a valid multivariate CDF—a task with more structured constraints that are easier to enforce than PDF validity.

We instantiate this CDF-first framework by modeling univariate conditional CDFs using Smooth Min-Max (SMM) networks—infinitely differentiable modules whose outputs, after appropriate normalization, satisfy CDF properties. For multivariate outputs, we adopt an autoregressive decomposition:
\begin{equation}
    p(\mathbf{y} \mid \mathbf{x}) = \prod_{i=1}^{d_y} p(y_i \mid \mathbf{x}, \mathbf{y}_{<i}),
\end{equation}
where each conditional PDF factor \(p(y_i \mid \mathbf{x}, \mathbf{y}_{<i})\) is obtained as the derivative of a corresponding conditional CDF \(F_i(y_i \mid \mathbf{x}, \mathbf{y}_{<i})\) modeled by an SMM network. This decomposition reduces multivariate CDE to a sequence of univariate CDF estimation tasks while preserving the full joint distribution. The resulting architecture guarantees PDF validity by construction, supports exact likelihood training via finite-difference differentiation, and preserves complex distributional shapes—all without discretization, kernel assumptions, or invertible maps. As shown in Figure~\ref{fig:structural_fidelity}, our method accurately recovers ground-truth distributions with sharp boundaries and topological holes where existing approaches fail. Our contributions are threefold:

\begin{itemize}
    \item We introduce a CDF-first framework for free-form CDE that reframes density estimation as learning a valid CDF—circumventing the ill-posedness of direct PDF estimation and minimizing inductive bias.

    \item We extend the framework to multivariate outputs via an autoregressive decomposition with SMM-based conditional CDFs, enabling tractable modeling of high-dimensional distributions while maintaining structural fidelity.

    \item  We demonstrate superior performance on both toy and real-world benchmarks, showing enhanced ability to capture multi-modality, skewness, and topological complexity compared to existing density estimation methods
\end{itemize}

The remainder of this paper is organized as follows: Section~\ref{section2} reviews related approaches; Section~\ref{section3} formalizes our CDF-first framework and describes the SMM architecture; Section~\ref{section4} presents experimental results; and Section~\ref{section5} discusses limitations and future directions.

\section{Related Work}\label{section2}
Our work addresses the fundamental challenge of free-form density estimation CDE by shifting from PDF estimation to CDF estimation. We review existing approaches through the lens of the inductive biases they impose to stabilize the ill-posed problem of density estimation—biases that our CDF-first framework avoids.

\subsection{Parametric and Flow-Based Density Estimation}
Direct PDF estimation methods impose strong structural assumptions. Mixture Density Networks (MDNs) \cite{citation:MDN} combine neural networks with parametric mixture models (typically Gaussians), introducing bias through a fixed kernel family that cannot adapt to arbitrary shapes, such as sharp boundaries or heavy tails.  Extensions such as Neural Autoregressive Density Estimators (NADE) \cite{pmlr-v15-larochelle11a} improve tractability but retain restrictive parametric forms. Normalizing flows \cite{citation:NF, citation:NVP} enhance flexibility via invertible transformations of a base density but are constrained by bijectivity. This invertibility requirement limits architectural design and, more fundamentally, precludes the modeling of distributions with topologically complex support (e.g., those with holes or disconnected components) \cite{cornish2021}, as seen in the failure of flow-based methods to preserve the central hole in the ring density in Figure~\ref{fig:structural_fidelity}. While improvements such as Neural Spline Flows (NSF) \cite{durkan2019neuralsplineflows} offer smoother transformations and Continuous Normalizing Flows (CNF) \cite{köhler2021smoothnormalizingflows} enable more flexible dynamics, they still inherit this topological limitation. Variational Autoencoders (VAEs) \cite{kingma2022autoencodingvariationalbayes} and their conditional variants \cite{NIPS2015_8d55a249} provide implicit density models but suffer from blurry approximations and challenging likelihood estimation. All these methods model the PDF directly, facing the inherent ill-posedness of differentiating finite-sample distributions.

\subsection{Discretization and Histogram-Based Methods}
To avoid parametric forms, some methods discretize the output space and model a piecewise-constant density. PixelCNN \cite{citation:pixelcnn} and Masked Autoregressive Flows (MAF) \cite{maskedautoregressiveflowdensity} learn conditional categorical distributions over bins, achieving impressive generative results but suffering from quantization artifacts, non-smooth densities, and poor scalability in high dimensions due to the curse of dimensionality. Methods like JBCENN \cite{LI_2021} improve bin efficiency, and the Deconvolutional Density Network (DDN) \cite{Chen2021DeconvolutionalDN} mitigates spikiness through hierarchical smoothing, yet they remain fundamentally tied to a discretized representation. This binning bias often leads to overly smooth estimates that blur sharp structural details (Figure~\ref{fig:structural_fidelity}, DDN), limiting the recovery of fine-grained density features.

\subsection{CDF and Quantile-Based Modeling}
Modeling the distribution function directly is a less explored but theoretically stable alternative. Quantile Regression Forests \cite{10.5555/1248547.1248582} estimate conditional quantiles nonparametrically but lack differentiability and tractable likelihoods. Implicit Quantile Networks (IQN)~\cite{dabney2018implicitquantilenetworksdistributional} learn quantile functions through implicit representations but do not provide explicit PDFs or exact log-likelihoods. Copula-based models \cite{NIPS2010_2a79ea27} separate marginal distributions from dependence structures but typically rely on parametric copula families. Monotonic networks \cite{NIPS1997_83adc922} enforce monotonicity constraints on the CDF, but these models are primarily focused on univariate outputs and do not generalize to multivariate conditional density estimation.

The most relevant work to ours is the Smooth Min-Max (SMM) network \cite{Igel_2023}, which provides a compact, infinitely differentiable parameterization for monotonic functions. SMM was originally proposed for modeling monotonic functions in regression tasks, but we repurposed it as a general-purpose CDF approximator within a framework for multivariate density estimation. Crucially, our work is the first to take advantage of the well-posedness of CDF estimation as a principled solution to the instability of free-form CDE, using only the definition link \( p(y|x) = \partial_y F(y|x) \) without commitments to kernels, invertibility, or bins. As shown in Figure~\ref{fig:structural_fidelity} (CDF-First column), this approach uniquely preserves complex topological features while maintaining smoothness.

In summary, existing CDE methods regularize the ill-posed problem of PDF estimation through significant inductive biases—fixed kernels, invertible maps, or discretization. These biases restrict the expressivity of the models, particularly when the true distribution deviates from the assumed structure. By theoretically establishing and practically instantiating a stable CDF-first paradigm, our framework minimizes these biases to enable true free-form density estimation. Our approach ensures stability, exact likelihood computation, and structural fidelity, addressing the limitations of prior methods in modeling complex conditional distributions.


\section{Methodology}\label{section3}
\begin{figure*}
    \centering
    \includegraphics[width=2\columnwidth]{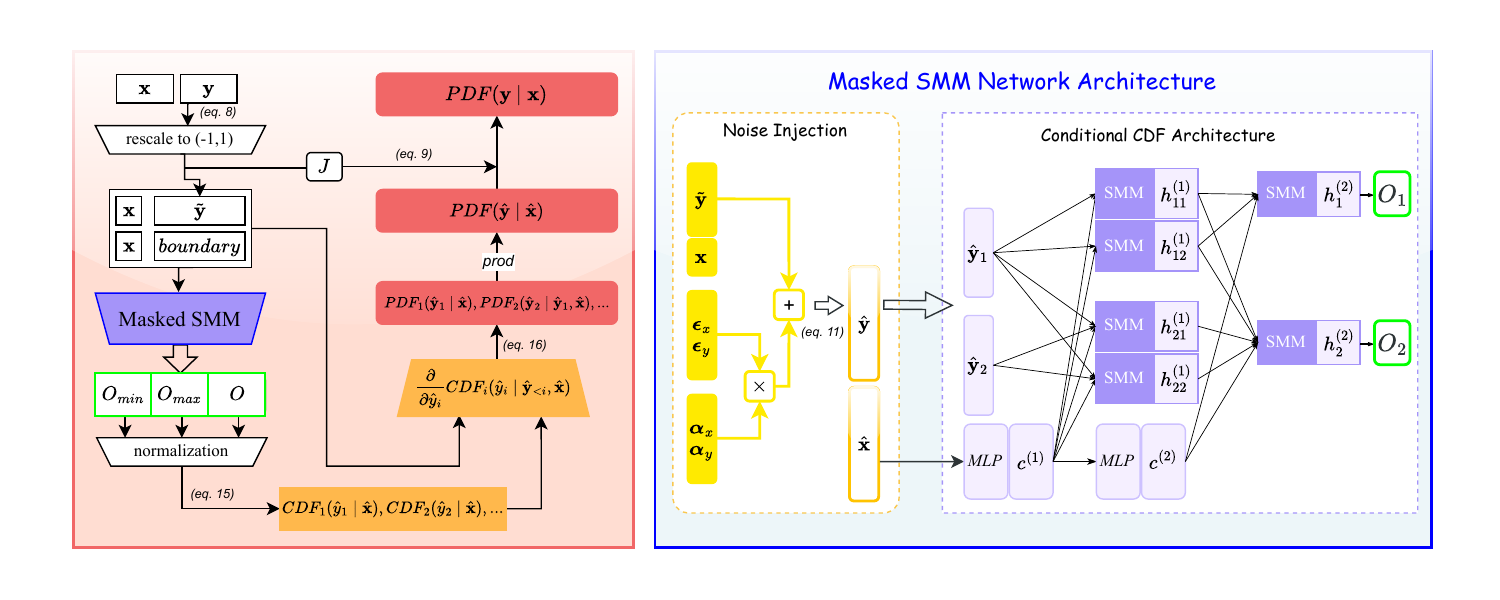}
    \caption{End-to-end architecture of the CDF-first framework: input normalization, stochastic noise injection, autoregressive masked SMM networks for CDF estimation, and boundary normalization with finite-difference PDF derivation.}
    \label{fig:architecture}
\end{figure*}

We propose a CDF-first framework for density estimation that models the conditional cumulative distribution function $F(\mathbf{y} \mid \mathbf{x})$ directly and derives the probability density $p(\mathbf{y} \mid \mathbf{x})$ via finite differences. The framework comprises four components: \textit{(i)} input data normalization, \textit{(ii)} stochastic regularization via noise injection, \textit{(iii)} autoregressive CDF estimation using masked Smooth Min-Max networks, and \textit{(iv)} boundary-normalized CDF parameterization with PDF derivation. Figure~\ref{fig:architecture} illustrates the complete workflow.

\subsection{Problem Formulation and Notation}\label{sec:problem_formulation}
Let $\mathbf{x} \in \mathcal{X} \subseteq \mathbb{R}^{d_x}$ denote a conditioning covariate and $\mathbf{y} \in \mathcal{Y} \subseteq \mathbb{R}^{d_y}$ a multivariate response. The goal is to model the conditional law $\mathbb{P}(\mathbf{y} \mid \mathbf{x})$. We adopt a CDF-first perspective by modeling the conditional cumulative distribution function:
\begin{equation}\label{eq:cdf_def}
    F(\mathbf{y} \mid \mathbf{x}) := \mathbb{P}(Y_1 \leq y_1, \dots, Y_{d_y} \leq y_{d_y} \mid \mathbf{x}),
\end{equation}
which uniquely determines $\mathbb{P}(\mathbf{y} \mid \mathbf{x})$.

For tractable multivariate modeling, we employ an autoregressive factorization of the joint probability density function:
\begin{equation}\label{eq:pdf_factorization}
    p(\mathbf{y} \mid \mathbf{x}) = \prod_{i=1}^{d_y} p(y_i \mid \mathbf{x}, \mathbf{y}_{<i}),
\end{equation}
where $\mathbf{y}_{<i} := (y_1, \dots, y_{i-1})$. Each conditional PDF factor $p(y_i \mid \mathbf{x}, \mathbf{y}_{<i})$ is obtained as the derivative of a corresponding conditional CDF:
\begin{equation}\label{eq:cdf_chain}
    p(y_i \mid \mathbf{x}, \mathbf{y}_{<i}) = \frac{\partial}{\partial y_i} F_i(y_i \mid \mathbf{x}, \mathbf{y}_{<i}),
\end{equation}
where $F_i(y_i \mid \mathbf{x}, \mathbf{y}_{<i}) := \mathbb{P}(Y_i \leq y_i \mid \mathbf{x}, \mathbf{Y}_{<i} = \mathbf{y}_{<i})$.

Thus, the joint PDF can be expressed as:
\begin{equation}\label{eq:joint_pdf_from_cdfs}
    p(\mathbf{y} \mid \mathbf{x}) = \prod_{i=1}^{d_y} \frac{\partial}{\partial y_i} F_i(y_i \mid \mathbf{x}, \mathbf{y}_{<i}).
\end{equation}
This formulation reduces multivariate conditional density estimation to a sequence of univariate conditional CDF estimation tasks. By learning smooth CDFs $F_i$ that are non-decreasing and satisfy $F_i(y_i \mid \mathbf{x}, \mathbf{y}_{<i}) \to 0$ as $y_i \to -\infty$ and $F_i(y_i \mid \mathbf{x}, \mathbf{y}_{<i}) \to 1$ as $y_i \to +\infty$, we guarantee that the resulting PDF is non-negative and integrates to one.



\subsection{Input Data Normalization}
\label{subsec:normalization}
To ensure numerical stability when using Smooth Min-Max networks, we normalize each dimension of the response vector $\mathbf{y}$ to $[-1, 1]$. Given training data $\mathcal{D} = \{(\mathbf{x}^{(n)}, \mathbf{y}^{(n)})\}_{n=1}^N$, we compute dimension-wise statistics:
\[
    y_{\min}^{(j)} = \min_{n} y_j^{(n)}, \quad y_{\max}^{(j)} = \max_{n} y_j^{(n)}, \quad j = 1, \dots, d_y.
\]

The normalization mapping $\Psi: \mathbb{R}^{d_y} \to [-1, 1]^{d_y}$ is defined as:
\begin{equation}\label{eq:normalize_target}
    \tilde{\mathbf{y}} = \Psi(\mathbf{y}) = 2 \frac{\mathbf{y} - \mathbf{y}_{\min}}{\mathbf{y}_{\max} - \mathbf{y}_{\min}} - 1,
\end{equation}

For conditioning covariates $\mathbf{x}$, we rely on architectural constraints rather than explicit normalization. The condition embedding network uses $\tanh$ activations, naturally bounding embeddings to $(-1, 1)$, ensuring numerical alignment with normalized responses.

During inference, densities on the normalized space are transformed back via the change-of-variables formula:
\begin{equation}\label{eq:jacobian}
    p(\mathbf{y} \mid \mathbf{x}) = p(\tilde{\mathbf{y}} \mid \mathbf{x}) \cdot \left| \det J_{\Psi}(\mathbf{y}) \right| = p(\tilde{\mathbf{y}} \mid \mathbf{x}) \cdot \prod_{j=1}^{d_y} c_j,
\end{equation}
where \(c_j = \frac{2}{y_{\max}^{(j)} - y_{\min}^{(j)}}\). $J_{\Psi}(\mathbf{y})$ is the diagonal Jacobian of $\Psi$ at $\mathbf{y}$.

\subsection{CDF-First Estimation with Masked SMM Networks}
\label{subsec:val_smm}
This section details the core architecture for learning conditional CDFs $F_i(y_i \mid \mathbf{x}, \mathbf{y}_{<i})$ using Smooth Min-Max networks. The SMM architecture provides smooth, bounded approximations to min-max compositions, naturally aligning with CDF properties.

\subsubsection{Regularization via Noise Injection}\label{subsubsec:noise}
To mitigate overfitting, we inject additive Gaussian noise during training \cite{citation:vae}. Given normalized responses $\tilde{\mathbf{y}}$ and covariates $\mathbf{x}$, we sample:
\[
\boldsymbol{\epsilon}_x \sim \mathcal{N}(\mathbf{0}, \mathbf{I}_{d_x}), \quad \boldsymbol{\epsilon}_y \sim \mathcal{N}(\mathbf{0}, \mathbf{I}_{d_y}),
\]
and create perturbed inputs:
\begin{equation}\label{eq:noise_injection}
    \hat{\mathbf{x}} = \mathbf{x} + \boldsymbol{\alpha}_x \odot \boldsymbol{\epsilon}_x, \quad \hat{\mathbf{y}} = \tilde{\mathbf{y}} + \boldsymbol{\alpha}_y \odot \boldsymbol{\epsilon}_y,
\end{equation}
where $\boldsymbol{\alpha}_x \in \mathbb{R}^{d_x}_+$, $\boldsymbol{\alpha}_y \in \mathbb{R}^{d_y}_+$ are learnable scaling parameters. This operation convolves the target distribution with a Gaussian kernel, preventing collapse onto discrete data points. For each response $\tilde{\mathbf{y}}$ and its associated boundary variables, we reuse the same
noise realization $\boldsymbol{\epsilon}_y$ to perturb them, so that the relative ordering
(monotonicity) between $\hat{\mathbf{y}}$ and $\tilde{\mathbf{y}}$ within each pair is preserved.

\subsubsection{Conditional CDF Architecture}\label{subsubsec:core_smm}
\textbf{Smooth Function Approximation.}
Following the Smooth Monotonic Model (SMM), we utilize soft extremum operations to approximate general monotonic functions. Let $\mathbf{h}^{(l)} \in \mathbb{R}^{H}$ denote hidden features at layer $l$:
\begin{equation}
\mathbf{h}^{(l)} = \text{SoftMin}_k \left( \text{SoftMax}_j \left( \mathbf{h}^{(l)}_{\text{pre}} \right)_{j \in \mathcal{G}_k} \right),
\end{equation}
where $\mathcal{G}_k$ is the $k$-th feature group. As SoftMax and SoftMin are smooth, strictly increasing approximations of the max and min operators, the output $\mathbf{h}^{(l)}$ remains strictly monotonically increasing with respect to the pre-activation input $\mathbf{h}^{(l)}_{\text{pre}}$. This design ensures the requisite monotonicity for CDF estimation while maintaining infinite differentiability.

\textbf{Conditional Feature Integration.} 
The condition $\hat{\mathbf{x}}$ is processed through layer-specific networks to extract context features:
\begin{equation}
    \begin{aligned}
    \mathbf{c}^{(1)} = \tanh\left(\text{MLP}^{(1)}({\hat{\mathbf{x}}})\right) \in (-1, 1)^{H_c}, \\
    \mathbf{c}^{(l+1)} = \tanh\left(\text{MLP}^{(l)}({\mathbf{c}^{(l)}})\right) \in (-1, 1)^{H_c},
    \end{aligned} 
\end{equation}
where $\text{MLP}^{(l)}$ projects the condition into the latent feature space. These conditions are integrated via additive modulation:
\begin{equation}\label{eq:masked_smm}
\begin{aligned}
    \mathbf{h}^1_{\text{pre}} &= \exp(\mathbf{W}_z^{(1)}) \mathbf{\hat{y}} + \mathbf{W}_c^{(1)} \mathbf{c}^{(1)} + \mathbf{b}^{(1)}, \\
    \mathbf{h}^{(l+1)}_{\text{pre}} &= \exp(\mathbf{W}_z^{(l)}) \mathbf{h}^{(l)} + \mathbf{W}_c^{(l)} \mathbf{c}^{(l)} + \mathbf{b}^{(l)}.
\end{aligned}
\end{equation}
\textbf{Structural Monotonicity and Ordinal Preservation.} 
The state transition weights are parameterized as $\exp(\mathbf{W}_z^{(l)})$, ensuring strict positivity. Consequently, the linear mapping from $\mathbf{h}^{(l)}$ to $\mathbf{h}^{(l+1)}_{\text{pre}}$ is strictly order-preserving, with the conditional terms $\mathbf{W}_c^{(l)} \mathbf{c}^{(l)}$ acting merely as instance-specific biases.
By composing these positive linear transformations with the monotonic SMM activations, the network constitutes a globally monotonic function $\Phi(\cdot)$ with respect to its input $\hat{\mathbf{y}}$.

Combiniing this with the synchronized noise injection strategy (Sec.~\ref{subsubsec:noise}), we establish an end-to-end ordinal guarantee. Since the same noise realization is applied to both the response $\tilde{\mathbf{y}}$ and its class boundaries, their relative order is maintained in the perturbed space. Because the network $\Phi$ is strictly monotonically increasing, this order is preserved in the latent feature space. 
Thus, for every batch instance, the latent representation of the response $\mathbf{h}(\hat{\mathbf{y}})$ remains strictly bounded by the latent representations of its corresponding boundaries, ensuring logical consistency throughout the training process.

\textbf{Autoregressive Masking} To enforce the autoregressive dependency $F_i(y_i \mid \mathbf{x}, \mathbf{y}_{<i})$, we apply binary masks similar to MADE~\cite{MADE} to input connections. For dimension $i$, define:
\[
\mathbf{M}^{(i)}_{j} = \begin{cases}
1 & \text{if } j < i \\
0 & \text{otherwise}
\end{cases}.
\]
This ensures $\partial F_i / \partial \hat{y}_j = 0$ for all $j > i$, implementing the desired causal structure while permitting simultaneous computation of all $d_y$ conditional CDFs.

\subsection{From CDF to PDF: Derivation via Finite Differences}\label{subsec:pdf_derivation}
The SMM network produces raw outputs $O(\hat{y}_i \mid \hat{\mathbf{x}}, \hat{\mathbf{y}}_{<i})$. To obtain valid CDFs satisfying boundary conditions, we apply linear rescaling.

\textbf{Boundary Normalization} For each dimension $i$, evaluate:
\[
O_{\min,i} = O(-1 \mid \hat{\mathbf{x}}, \hat{\mathbf{y}}_{<i}), \quad O_{\max,i} = O(1 \mid \hat{\mathbf{x}}, \hat{\mathbf{y}}_{<i}).
\]
The normalized CDF is:
\begin{equation}\label{eq:boundary_norm}
    \hat{F}_i(\hat{y}_i \mid \hat{\mathbf{x}}, \hat{\mathbf{y}}_{<i}) = \frac{O(\hat{y}_i \mid \hat{\mathbf{x}}, \hat{\mathbf{y}}_{<i}) - O_{\min,i}}{O_{\max,i} - O_{\min,i}} \in [0, 1].
\end{equation}
This guarantees $\hat{F}_i(-1) = 0$ and $\hat{F}_i(1) = 1$ while preserving monotonicity and differentiability.

\textbf{PDF Recovery via Finite Differences} The conditional PDF is obtained by differentiating the CDF. We approximate the derivative using central finite differences:
\begin{equation}\label{eq:PDF_derivation}
    \hat{p}(\hat{y}_i \mid \hat{\mathbf{x}}, \hat{\mathbf{y}}_{<i}) \approx \frac{\hat{F}_i(\hat{y}_i + \delta \mid \hat{\mathbf{x}}, \hat{\mathbf{y}}_{<i}) - \hat{F}_i(\hat{y}_i - \delta \mid \hat{\mathbf{x}}, \hat{\mathbf{y}}_{<i})}{2\delta},
\end{equation}
where $\delta = 5 \times 10^{-6}$ (analyzed in Section~\ref{paper:hyperparameter_delta}). This avoids second-order automatic differentiation while maintaining accuracy for smooth functions. During inference, we set $\hat{\mathbf{x}} = \mathbf{x}$ and $\hat{\mathbf{y}} = \tilde{\mathbf{y}}$, then apply the change-of-variables formula \eqref{eq:jacobian} to obtain $p(\mathbf{y} \mid \mathbf{x})$.

\begin{algorithm}[H]
\caption{Training Procedure for CDF-First Density Estimation}
\label{alg:training}
\begin{algorithmic}[1]
\REQUIRE Training dataset $\mathcal{D} = \{(\mathbf{x}^{(n)}, \mathbf{y}^{(n)})\}_{n=1}^N$
\REQUIRE Hyperparameters: $\eta$ (learning rate), $B$ (batch size), $\beta_x, \beta_y$ (reg. strengths), $\delta$ (finite diff. step)
\REQUIRE Precomputed normalization statistics: $\mathbf{y}_{\min}, \mathbf{y}_{\max}$
\ENSURE Trained parameters $\Theta$ (SMM network), $\boldsymbol{\alpha}_x, \boldsymbol{\alpha}_y$ (noise scales)

\STATE Initialize $\Theta$, $\boldsymbol{\alpha}_x$, $\boldsymbol{\alpha}_y$
\FOR{epoch $=1$ to $E$}
    \FOR{each minibatch $\{(\mathbf{x}^{(b)}, \mathbf{y}^{(b)})\}_{b=1}^B \subset \mathcal{D}$}
        \STATE $\tilde{\mathbf{y}}^{(b)} \gets \Psi(\mathbf{y}^{(b)})$ \COMMENT{Normalize via \eqref{eq:normalize_target}}
        \STATE Sample $\boldsymbol{\epsilon}_x, \boldsymbol{\epsilon}_y \sim \mathcal{N}(\mathbf{0}, \mathbf{I})$
        \STATE $\hat{\mathbf{x}}^{(b)} \gets \mathbf{x}^{(b)} + \boldsymbol{\alpha}_x \odot \boldsymbol{\epsilon}_x$ \COMMENT{Noise injection via \eqref{eq:noise_injection}}
        \STATE $\hat{\mathbf{y}}^{(b)} \gets \tilde{\mathbf{y}}^{(b)} + \boldsymbol{\alpha}_y \odot \boldsymbol{\epsilon}_y$
        \STATE Compute $O(\hat{y}_i^{(b)} \mid \hat{\mathbf{x}}^{(b)}, \hat{\mathbf{y}}_{<i}^{(b)})$ via SMM network
        \STATE Compute $\hat{F}_i$ via boundary normalization \eqref{eq:boundary_norm}
        \STATE Compute $\hat{p}$ via finite differences \eqref{eq:PDF_derivation}
        \STATE Transform to original space via \eqref{eq:jacobian}
        \STATE Compute $\mathcal{L}_{\text{NLL}}$ via \eqref{eq:loss_nll}
        \STATE Compute $\mathcal{L}_{\text{Reg}}$ via \eqref{eq:loss_kl}
        \STATE $\mathcal{L} \gets \mathcal{L}_{\text{NLL}} + \mathcal{L}_{\text{Reg}}$
        \STATE Update parameters via gradient descent on $\mathcal{L}$
    \ENDFOR
\ENDFOR 
\STATE \textbf{return} $\Theta, \boldsymbol{\alpha}_x, \boldsymbol{\alpha}_y$
\end{algorithmic}
\end{algorithm}

\subsection{Training Objective and Loss Function}\label{subsec:loss}
The model is trained by minimizing a composite loss:
\begin{equation} \label{eq:total_loss}
    \mathcal{L} = \mathcal{L}_{\text{NLL}} + \mathcal{L}_{\text{Reg}}.
\end{equation}

\textbf{Negative Log-Likelihood} The primary loss is the negative log-likelihood:
\begin{equation}\label{eq:loss_nll}
    \mathcal{L}_{\text{NLL}} = -\frac{1}{N} \sum_{n=1}^N \sum_{i=1}^{d_y} \log \hat{p}(y_i^{(n)} \mid \mathbf{x}^{(n)}, \mathbf{y}_{<i}^{(n)}),
\end{equation}
where each $\hat{p}(y_i \mid \mathbf{x}, \mathbf{y}_{<i})$ is obtained via equations \eqref{eq:PDF_derivation} and \eqref{eq:jacobian}.

\textbf{Variational Regularization.} To regulate noise magnitude, we optimize the following KL-divergence penalty:

\begin{equation}\label{eq:loss_kl}
\begin{aligned}
    \mathcal{L}_{\text{Reg}} ={}& \beta_x D_{\text{KL}}\!\left(q(\boldsymbol{\epsilon}_x) \parallel \mathcal{N}(\mathbf{0}, \mathbf{I}_{d_x})\right) \\
    &+ \beta_y D_{\text{KL}}\!\left(q(\boldsymbol{\epsilon}_y) \parallel \mathcal{N}(\mathbf{0}, \mathbf{I}_{d_y})\right),
\end{aligned}
\end{equation}

where $q(\boldsymbol{\epsilon}) = \mathcal{N}(\mathbf{0}, \operatorname{diag}(\boldsymbol{\alpha}^2))$. We employ distinct coefficients $\beta_x$ and $\beta_y$ to decouple the regularization of covariates and responses, as noise injected into the conditioning features plays a structurally different role than noise applied to the target inputs. 

Crucially, while we initialize $\boldsymbol{\alpha}$ to small values to facilitate early learning, the standard normal prior acts as a necessary barrier against deterministic collapse. Without this regularization, the reconstruction objective would drive $\boldsymbol{\alpha} \to \mathbf{0}$; the KL term (specifically $-\log \boldsymbol{\alpha}^2$) counteracts this tendency, maintaining a non-zero noise floor to prevent the module from degenerating into an identity mapping.

\section{Experiments and Results}\label{section4}

\subsection{Experimental Settings}
\textbf{Toy \& Real World Tasks}. To evaluate our method's ability to model complex distributional shapes, we designed four two-dimensional toy tasks where the ground truth conditional density \(p(\mathbf{y} \mid x)\) is known and can be computed analytically. These tasks were designed to exhibit challenging characteristics: multimodality, non-Gaussianity, rotational dependence on inputs, and topological complexity. In all tasks, the conditioning variable is sampled uniformly: \(x \sim \mathcal{U}(-1, 1)\). The details of generating training samples are illustrated in Appendix~\ref{app:toy_tasks}. For each toy task, we generated 2,000 training samples and 2,000 test samples, with the conditioning variable \(x\) sampled uniformly from \([-1, 1]\).


To validate practical utility, we evaluated our method on seven real-world regression datasets from the UCI Machine Learning Repository~\cite{Dua:2019}. These datasets span diverse domains and vary in sample size, input dimensionality, and output dimensionality, as summarized in Table~\ref{table:dataset}. Implementation details and baseline models can be found in Appendix~\ref{app:implementation_details}.

\begin{table}[t]
  \centering
    \caption{Characteristics of UCI benchmark datasets. For each dataset, we report the number of samples, input features (\(d_x\)), and output dimensions (\(d_y\)).}
  \label{table:dataset}
  \begin{tabular}{@{}lcccr@{}}
    \toprule
    \textbf{Dataset} & \textbf{Samples} & \(d_x\) & \(d_y\) & \textbf{Domain} \\
    \midrule
    Fish     & 908   & 6  & 1 & Toxicology \\
    Concrete  & 1,030 & 8  & 1 & Civil Eng. \\
    Energy      & 768   & 8  & 2 & Building \\
    Parkinsons & 5,875 & 16 & 2 & Medical \\
    Temperature & 7,588 & 21 & 2 & Env. \\
    Air           & 8,891 & 10 & 3 & Env. \\
    Skillcraft & 3,338 & 15 & 4 & Gaming \\
    \bottomrule
  \end{tabular}
\end{table}

\subsection{Ablation Study}\label{subsec:ablation}

To validate the necessity of each component in our CDF-first framework, we conducted an ablation study on the Elastic Ring toy task using 1000 samples with a 50-50 train-test split. We compared four configurations: \textit{(1)} the full CDF-first model with noise injection and SMM networks, \textit{(2)} removal of noise injection, \textit{(3)} replacement of SMM with standard min-max (MM) networks, and \textit{(4)} replacement of SMM with monotonic MLPs. Early stopping followed the toy experiment protocol based on SSE improvement.

\begin{figure}[h]
    \centering
\includegraphics[width=\columnwidth]{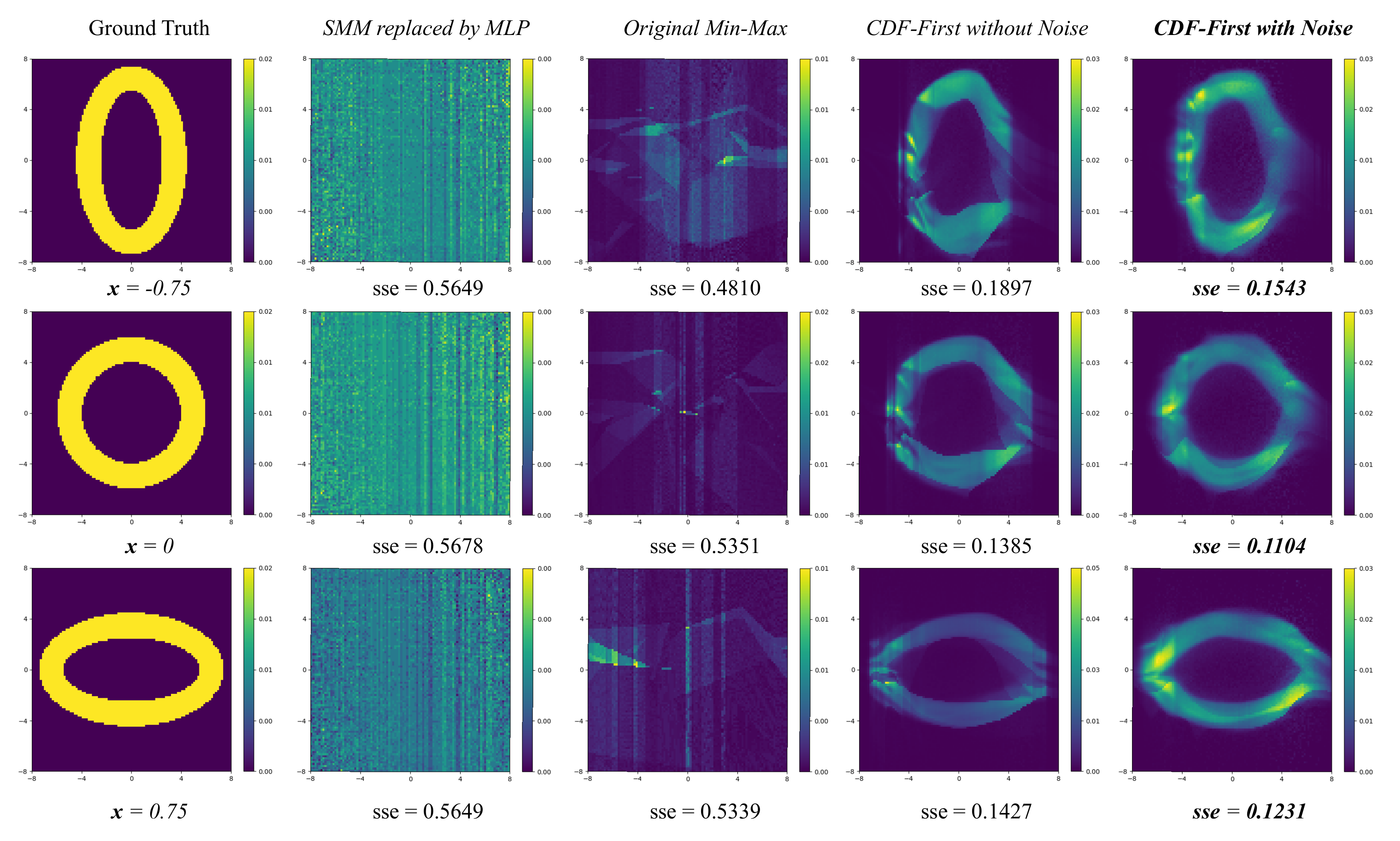}
    \caption{Ablation study on the Elastic Ring task. Each row shows conditional density estimates at $x = -0.75, 0, 0.75$ for a variant of our CDF-first framework, with sum-of-squared-errors (SSE) reported per column. Only the full model (bottom row) accurately recovers the ring’s topology; removing noise, replacing SMM with hard Min-Max, or using an MLP baseline all degrade structural fidelity, demonstrating the necessity of both noise injection and smooth SMM networks for accurate density estimation.}
    \label{fig:ablation_study}
\end{figure}

Figure~\ref{fig:ablation_study} presents the quantitative results across three conditioning values ($x = -0.75, 0, 0.75$), measured by sum of squared errors (SSE) between estimated and ground truth densities. The full CDF-first model achieves the lowest SSE across all conditions, with an average SSE of 0.1293 compared to 0.5659 for the MLP variant (77.2\% higher), 0.5167 for standard MM networks (75.0\% higher), and 0.1570 for the noise-free SMM configuration (17.6\% higher).

The results demonstrate that each architectural component contributes significantly to model performance. Removing noise injection increases SSE by an average of 22.3\%, indicating that stochastic regularization is essential for preventing overfitting to sparse training data. Replacing SMM with standard MM networks degrades performance by 75.0\%, highlighting the importance of smooth, differentiable CDF approximations for stable PDF derivation via finite differences. The monotonic MLP variant performs worst (77.2\% higher SSE), confirming that SMM's structured min-max compositions provide superior expressivity for complex distribution shapes while maintaining CDF validity.

Qualitatively, only the full model preserves the ring's elliptical geometry and aspect ratio variation with $x$, while ablation variants exhibit density leakage, boundary artifacts, or topological distortion. These results validate our architectural choices: noise injection provides necessary regularization in low-data regimes, and SMM networks provide the smooth, bounded function class necessary for stable CDF estimation and subsequent PDF derivation via finite differences.

\subsection{Analysis for Hyperparameter (\(\delta\))}\label{paper:hyperparameter_delta}
We analyzed the sensitivity of our method to the finite-difference step size $\delta$ used for PDF derivation (Eq.~\ref{eq:PDF_derivation}). On the Elastic Ring task (Table~\ref{tab:delta_sensitivity_transposed}), we evaluated $\delta \in \{5\times10^{-7}, 1\times10^{-6}, 5\times10^{-6}, 1\times10^{-5}, 3\times10^{-5}\}$ at $x \in \{-0.75, -0.25, 0, 0.25, 0.75\}$ and report mean SSE. The optimal value is $\delta = 5\times10^{-6}$ (SSE = 0.0694), with performance degrading for both smaller (truncation error) and larger (discretization error) values. Qualitative density estimates for different $\delta$ are provided in Figure~\ref{fig:hyperparameter_delta} (Appendix~\ref{app:delta}), confirming that our method is robust to small variations around the chosen $\delta$.

\begin{table}[h]
\centering
\caption{Sensitivity of finite-difference step size $\delta$ on the Elastic Ring task (SSE, lower is better).}
\label{tab:delta_sensitivity_transposed}
\begin{tabular}{lccccc}
\toprule
$\delta$ & 5e-7 & 1e-6 & 5e-6 & 1e-5 & 3e-5 \\
\midrule
SSE & 0.134 & 0.113 & \textbf{0.088} & 0.090 & 0.101 \\
\bottomrule
\end{tabular}
\end{table}

\begin{figure}[h]
    \centering
\includegraphics[width=\linewidth]{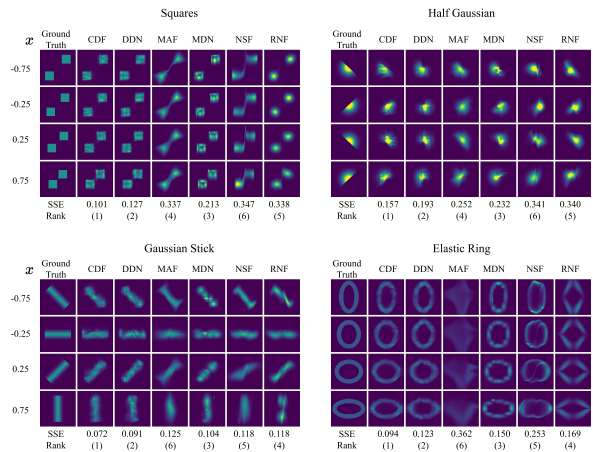}
    \caption{Conditional density estimation on four toy tasks (Squares, Half Gaussian, Gaussian Stick, Elastic Ring) at \(x = \{-0.75, -0.25, 0.25, 0.75\}\). Our CDF-First model consistently recovers ground-truth structures, including disconnected modes, sharp boundaries, and topological holes, outperforming all baselines (DDN, MAF, MDN, NSF, RNF) in both visual fidelity and SSE.}
    \label{fig:toy_performance}
\end{figure}
\subsection{Performance Analysis on Toy Tasks}\label{subsec:toy_results}
We evaluated all methods on four challenging 2D toy tasks—Disconnected Squares, Half-Gaussian, Gaussian Stick, and Elastic Ring—each designed to test specific aspects of distributional complexity: multimodality, non-Gaussianity, anisotropic covariance, and topological holes. For each task, we computed the sum of squared errors (SSE) between estimated and ground truth densities at four conditioning values \(x = \{-0.75, -0.25, 0.25, 0.75\}\), ranking methods by average SSE.

\begin{table*}[h]
  \centering
   \caption{Test negative log-likelihood (mean ± std, lower is better) on UCI datasets with output dimensionalities: Fish (1D) Concrete (1D), Energy (2D), Parkinsons (2D), Temperature (2D), Air (3D), Skillcraft (4D). Best results per dataset in bold.}
  \label{tab:real_tasks_performance}
  \begin{tabular}{cccccccc}
    \toprule
     & Fish & Concrete & Energy & Parkinsons & Temperature & Air & Skillcraft\\
    \midrule
    MDN & 0.92±0.03 & 0.76±0.09 & -0.99±0.15 & 1.26±0.05 &  1.08±0.03 & 1.26±0.04 & 2.27±1.14 \\
    MAF & 1.10±0.09 & 1.10±0.09 & -0.13±0.29 & 1.54±0.08 & 1.25±0.04 & 1.14±0.08 &  4.06±0.28  \\
    NSF & 1.57±0.19 & 1.32±0.29 & 9.23±1.63 & 1.72±0.13 & 1.79±0.16 & 1.50±0.17 & 6.72±0.64 \\
    RNF & 1.24±0.08 &  1.24±0.09 & 1.89±0.20 & 2.30±0.07 & 1.84±0.10 & 1.70±0.11 & 4.49±0.19 \\
    DDN & 0.92±0.04 & 0.59±0.03 & -1.66±0.18 & 0.36±0.04 & 0.74±0.04 & 0.93±0.05 & 1.66±0.07\\
    \midrule
    CDF-First  &  \textbf{0.89±0.01} &  \textbf{0.44±0.05} &  \textbf{-2.03±0.19} & 0.73±0.04 & \textbf{0.70±0.06} & 1.06±0.03 & \textbf{1.65±0.08} \\
    \bottomrule
  \end{tabular}
\end{table*}
As shown in Figure~\ref{fig:toy_performance}, CDF-First achieves the lowest SSE and highest rank (Rank 1) on all four tasks, with mean SSEs of 0.101 (Squares), 0.157 (Half Gaussian), 0.072 (Gaussian Stick), and 0.094 (Elastic Ring). The second-best method, DDN, consistently attains Rank 2 but exhibits 26–72\% higher SSE across tasks. Flow-based models (MAF, NSF, RNF) and MDN perform substantially worse, particularly on tasks with topological complexity; for example, on Elastic Ring, NSF achieves an SSE of 0.253—2.7 times higher than our method (0.094).

Only the proposed CDF-First method accurately recovers the defining structures of each task: the disconnected square modes without bridging the gap, the sharp half-Gaussian boundary without oversmoothing, the rotating anisotropic covariance of the Gaussian Stick, and the elliptical ring with varying aspect ratio while preserving the hollow interior. In contrast, MDN oversmooths multimodality, flow-based methods (MAF/NSF/RNF) fill topological holes or disjoint geometry, and even the strongest baseline (DDN) blurs sharp edges. These failures stem from their reliance on architectural assumptions—parametric mixtures, invertible transformations, or discretization—that are incompatible with free-form conditional density structures.

\subsection{Performance on Real-World Tasks}
We evaluated all methods on seven real-world datasets from the UCI repository, spanning output dimensionalities from 1 to 4. Table~\ref{tab:real_tasks_performance} reports negative log-likelihood (NNL; lower is better) averaged over \textit{ten-fold} (mean \(\pm\) standard deviation). CDF-First achieves the best NLL on five of the seven datasets (Fish, Concrete, Energy, Temparature, and Skillcraft), demonstrating consistent performance across varying output dimensions and sample sizes.

Notably, on the two-dimensional Energy dataset, CDF-First achieves an NLL of $-2.03$—a 22.3\% improvement over the second-best method (DDN, $-1.66$). On Concrete (1D), CDF-First improves NLL by 25.4\% over DDN (0.44 vs. 0.59). For the highest-dimensional task (Skillcraft, 4D), CDF-First maintains competitive performance with the best baseline while avoiding the instability seen in flow-based methods (NSF: 6.72, RNF: 4.49). The two exceptions are Parkinsons and Air, where DDN achieves slightly better NLL, though within one standard deviation in the case of Air (0.93 vs. 1.06).

These results demonstrate that our CDF-First framework maintains robust performance across diverse real-world conditions, from low to moderate output dimensionalities, without requiring task-specific architectural adjustments. The consistent improvements across most datasets highlight the advantage of direct CDF modeling over methods reliant on parametric mixtures (MDN), invertible flows (MAF, NSF, RNF), or discretization (DDN), particularly when distributions exhibit non-Gaussian characteristics not captured by standard parametric forms.

\begin{figure}
    \centering
    \includegraphics[width=\linewidth]{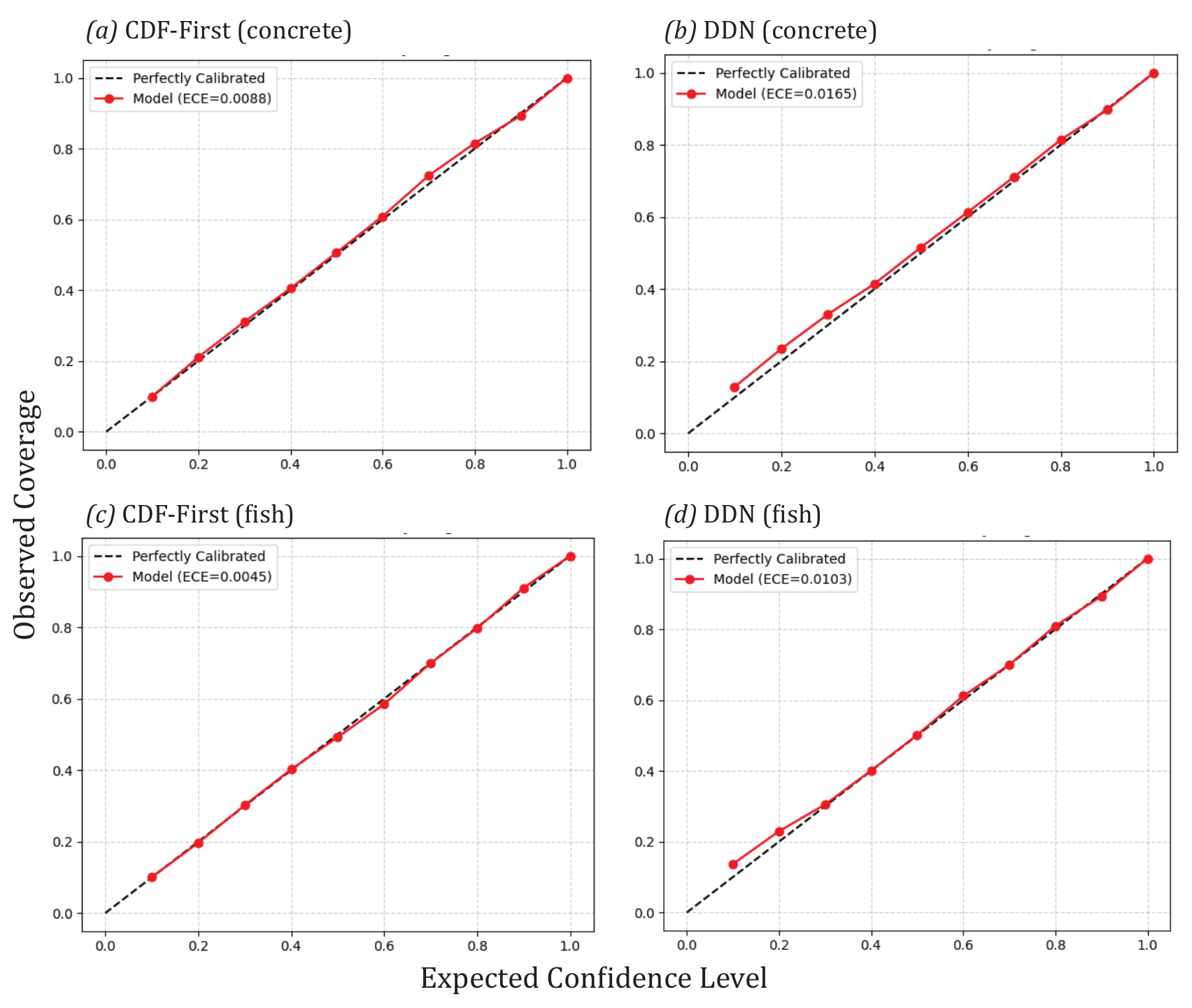}
    \caption{Reliability diagrams for CDF-First and DDN on Concrete (a,b) and Fish (c,d) datasets. CDF-First shows closer alignment to the diagonal (perfect calibration) and lower ECE values, indicating superior calibration performance.}
    \label{fig:ECE}
\end{figure}
\subsection{Calibration Analysis}
A primary goal of CDE is to produce well-calibrated uncertainty estimates. We evaluate calibration using the Expected Calibration Error (ECE) on two UCI datasets—Concrete and Fish—comparing our CDF-First model against DDN. CDF-First achieves significantly lower ECE: 0.0088 vs. 0.0165 on Concrete and 0.0045 vs. 0.0103 on Fish, demonstrating superior calibration. This confirms that modeling the CDF directly yields not only higher likelihoods but also more reliable probabilistic predictions. Detail analysis can be found in Appendix~\ref{app:calibration_analysis}.

\section{Conclusion}\label{section5}
We have presented a CDF‑first framework for conditional density estimation that directly models the cumulative distribution function and derives densities via finite differences. By learning smooth, valid CDFs using masked Smooth Min‑Max networks, our approach avoids the ill‑posedness of PDF estimation and imposes minimal inductive bias. 

The primary limitation of our work is the sequential sampling requirement of autoregressive factorization. Future work will explore parallel sampling techniques and extensions to higher-dimensional outputs. Nevertheless, by shifting the estimation target from PDF to CDF, we provide a stable, expressive foundation for probabilistic modeling that preserves structural fidelity without restrictive assumptions.

\section*{Impact Statements}
This paper presents work whose goal is to advance the field of Machine Learning. There are many potential societal consequences of our work, none which we feel must be specifically highlighted here.

\bibliography{example_paper}
\bibliographystyle{icml2026}

\newpage
\appendix
\onecolumn

\section{Implementation Details}\label{app:implementation_details}
\textbf{Baseline Models}. We compared our CDF-first method against five established conditional density estimation approaches. \textbf{MDN} (Mixture Density Network) used 20 Gaussian kernels with full covariance matrices. \textbf{MAF} (Masked Autoregressive Flow) employed 5 autoregressive transformations with linear spline coupling. \textbf{NSF} (Neural Spline Flow) utilized 5 autoregressive rational-quadratic spline transformations. \textbf{RNF} (Radial Normalizing Flow) implemented 5 radial flow transformations with elliptical basis functions. \textbf{DDN} (Discrete Density Network) used 256 bins per dimension with 16 latent codes, permutation order 1, and \(\beta=0.02\) for regularization. All baseline models were implemented using their standard open-source libraries and optimized under identical training conditions (hardware, optimizer, learning rate, and data splits) as our method to ensure fair comparison.

\textbf{Evaluation Metrics}. For real tasks, we evaluated models using negative log-likelihood (NLL) on held-out test data, measuring probability density estimation quality. For toy tasks where ground truth densities are analytically known, we computed the sum of squared errors (SSE) between estimated and true densities on a fixed evaluation grid.


For real-world UCI datasets, we used a 30\%/70\% train-test split with early stopping based on test NLL. Noise regularization was $\beta_x = \beta_y = 0.005$, with $\log\sigma$ initialized at -2. The SMM architecture used $[16,16,1]$ target features and $[8,8,2]$ condition features per layer, with batch normalization applied after each SMM layer.

All inputs and outputs were normalized to $[-1, 1]$ following Section~\ref{subsec:normalization}. PDFs were derived via finite differences with $\delta = 5 \times 10^{-6}$. We used three SMM layers with condition embedding MLPs of matching dimensions and $\tanh$ activations. The group structure remained 32 groups of size 32 (1024 total units) across all experiments.

\section{Toy Task Specifications}\label{app:toy_tasks}
\begin{figure}
    \centering
    \includegraphics[width=1\linewidth]{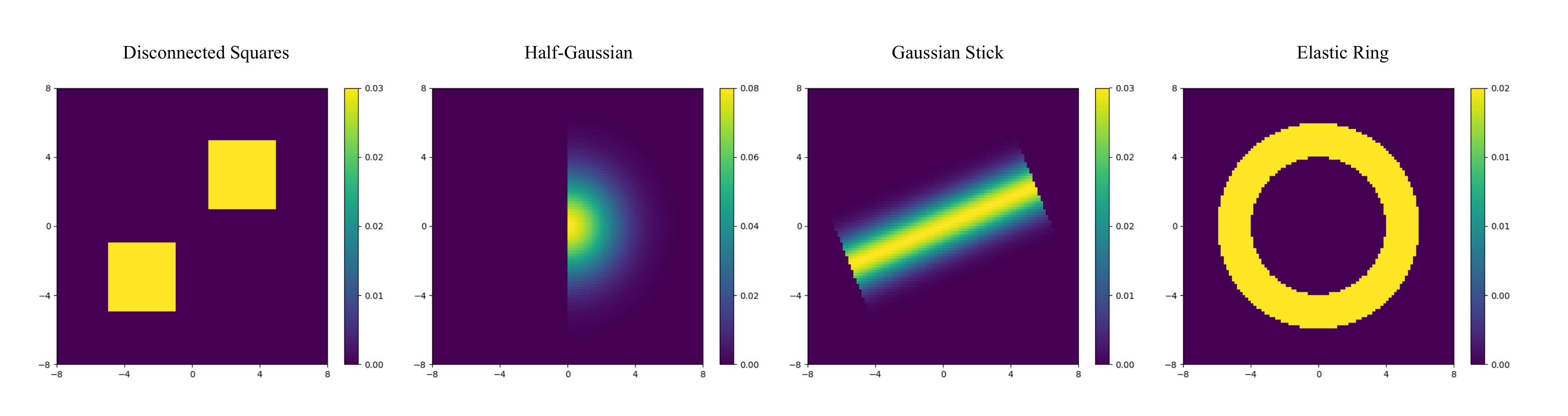}
    \caption{Four synthetic tasks used for evaluation: Disconnected Squares, Half-Gaussian, Gaussian Stick, and Elastic Ring. Each task is designed to test a specific challenge in conditional density estimation—multimodality, non-Gaussianity, rotational dependence, and topological complexity, respectively.}
    \label{fig:toy_appendix}
\end{figure}

To evaluate structural fidelity, we design four 2D synthetic conditional density estimation tasks with known ground-truth \( p(\mathbf{y} \mid x) \), where \( x \sim \mathcal{U}(-1, 1) \). Each task exhibits distinct challenges: multimodality, non-Gaussianity, anisotropic covariance, or topological complexity (see Figure~\ref{fig:toy_appendix}).
\begin{itemize}
    \item \textbf{Disconnected Squares:} This task features a bimodal distribution with two disconnected square supports whose positions shift linearly with \(x\). Given \(x\), we sample:
    \[
    \lambda \sim \mathrm{Bernoulli}(0.5), \quad 
    a^{(1)}, a^{(2)} \stackrel{\text{iid}}{\sim} \mathcal{U}(-5 + x, -1 + x), \quad 
    b^{(1)}, b^{(2)} \stackrel{\text{iid}}{\sim} \mathcal{U}(1 - x, 5 - x),
    \]
    and define the response as:
    \[
    y^{(1)} = \lambda a^{(1)} + (1 - \lambda) b^{(1)}, \quad 
    y^{(2)} = \lambda a^{(2)} + (1 - \lambda) b^{(2)}.
    \]
    The resulting conditional distribution has two equally probable modes with supports \([ -5+x, -1+x ]^2\) and \([ 1-x, 5-x ]^2\).

    \item \textbf{Half-Gaussian:} This task features a non-Gaussian, rotated distribution where one marginal follows a half-normal distribution. Given \(x\), we sample:
    \[
    a, b \stackrel{\text{iid}}{\sim} \mathcal{N}(0, 2),
    \]
    and apply a rotation by angle \(x\pi\):
    \[
    y^{(1)} = |a| \cos(x\pi) - b \sin(x\pi), \quad 
    y^{(2)} = |a| \sin(x\pi) + b \cos(x\pi).
    \]
    The distribution combines a half-normal component (\(|a|\)) with a normal component (\(b\)), rotated based on \(x\).

    \item \textbf{Gaussian Stick:} This task creates a distribution with highly anisotropic covariance structure that rotates with \(x\). Given \(x\), we sample:
    \[
    a \sim \mathcal{N}(0, 1), \quad b \sim \mathcal{U}(-6, 6),
    \]
    define the rotation angle \(c = (-0.75 + x)/2\), and apply:
    \[
    y^{(1)} = a \cos(c\pi) - b \sin(c\pi), \quad 
    y^{(2)} = a \sin(c\pi) + b \cos(c\pi).
    \]
    The resulting distribution has Gaussian spread in one direction and uniform spread in the orthogonal direction, with orientation depending on \(x\).

    \item \textbf{Elastic Ring:} This task creates a topologically complex distribution with a ring-shaped support whose aspect ratio varies with \(x\). Given \(x\), we sample:
    \[
    d \sim \mathcal{U}(0, 2), \quad \theta \sim \mathcal{U}(0, 2\pi),
    \]
    and define:
    \[
    y^{(1)} = (4 + 2x + d) \cos\theta, \quad 
    y^{(2)} = (4 - 2x + d) \sin\theta.
    \]
    The ring's elliptical shape changes with \(x\): for \(x = -1\), it is stretched along the x-axis; for \(x = 1\), along the y-axis.
\end{itemize}

\section{Extended Hyperparameter Analysis \(\delta\)} \label{app:delta}
\begin{figure*}
    \centering
    \includegraphics[width=\textwidth]{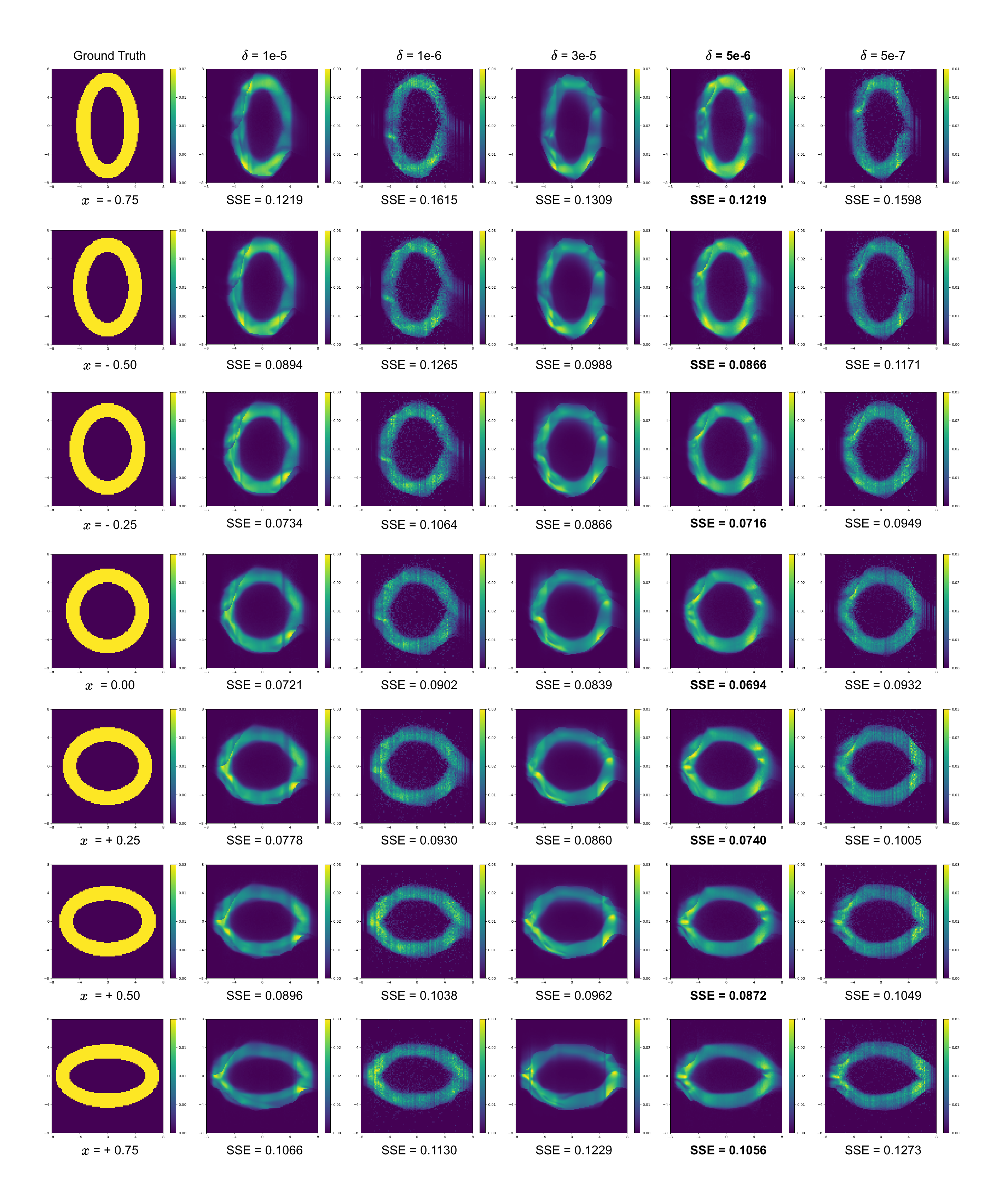}
    \caption{Conditional density estimates on the Elastic Ring task for different finite-difference step sizes $\delta$. Each column corresponds to a $\delta$ value, each row to a conditioning value $x$, with SSE values averaged over $x$. Optimal performance occurs at $\delta = 5\times10^{-6}$.}
    \label{fig:hyperparameter_delta}
\end{figure*}

We evaluate the sensitivity of the finite-difference approximation to the step size \(\delta\) on the Elastic Ring task across seven conditioning values \(x \in \{-0.75, -0.5, -0.25, 0, 0.25, 0.5, 0.75\}\). As shown in Figure~\ref{fig:hyperparameter_delta}, performance is optimal at \(\delta = 5 \times 10^{-6}\), achieving the lowest average SSE (0.0694). Larger values (e.g., \(1\times10^{-5}\)) introduce truncation error, while smaller values (e.g., \(5\times10^{-7}\)) amplify numerical noise—both degrading density fidelity. This confirms that \(\delta = 5\times10^{-6}\) balances approximation accuracy and floating-point stability.

\section{Extended Calibration Analysis}\label{app:calibration_analysis}
\begin{figure}
    \centering
    \includegraphics[width=0.6\linewidth]{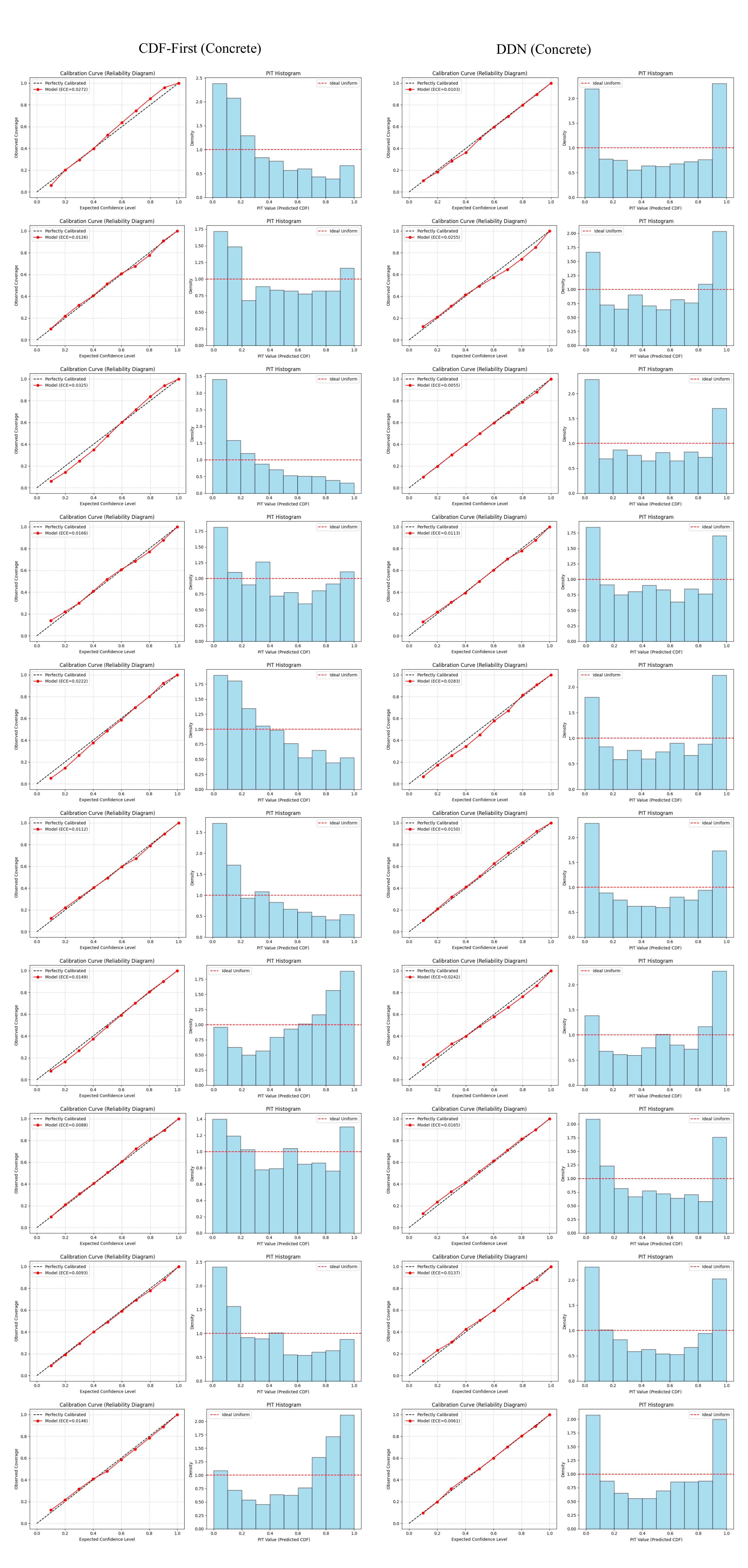}
        \caption{Reliability diagram for Concrete dataset (10 bins). CDF-First achieves ECE = 0.0088, compared to DDN's 0.0165, demonstrating superior calibration.}
    \label{fig:ECE_concrete}
\end{figure}

\begin{figure}
    \centering
    \includegraphics[width=0.6\linewidth]{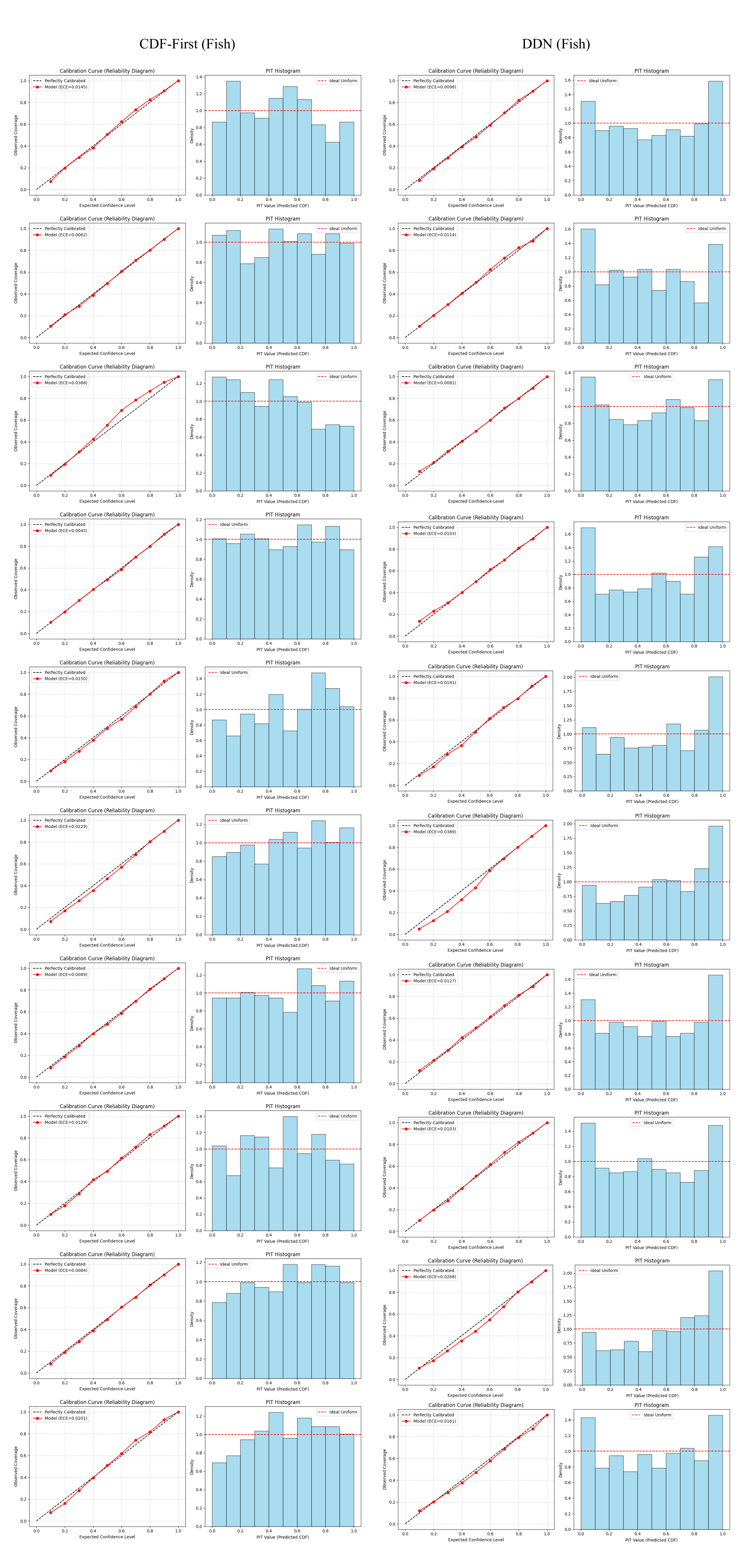}
    \caption{Reliability diagram for Fish dataset (10 bins). CDF-First achieves ECE = 0.0045, compared to DDN's 0.0103, showing better alignment with the diagonal (perfect calibration).}
    \label{fig:ECE_fish}
\end{figure}

We provide additional calibration analysis using 10-bin reliability diagrams for the Concrete and Fish datasets, comparing CDF-First with the strongest baseline (DDN). The results, shown in Figures~\ref{fig:ECE_concrete} and~\ref{fig:ECE_fish}, demonstrate consistent calibration improvements: CDF-First achieves ECE values of 0.0088 (Concrete) and 0.0045 (Fish), outperforming DDN (0.0165 and 0.0103, respectively). In both cases, CDF-First's reliability curve remains closer to the diagonal, indicating better-calibrated uncertainty estimates across all confidence levels.

\end{document}